\documentclass[conference]{IEEEtran}
\IEEEoverridecommandlockouts
\usepackage{cite}
\usepackage{amsmath,amssymb,amsfonts}
\usepackage{algorithmic}
\usepackage{graphicx}
\usepackage{textcomp}
\usepackage{xcolor}
\usepackage{amsfonts}
\usepackage{amssymb}
\usepackage{amsmath}
\usepackage{graphicx}
\usepackage{caption}
\usepackage{subcaption}
\def\BibTeX{{\rm B\kern-.05em{\sc i\kern-.025em b}\kern-.08em
    T\kern-.1667em\lower.7ex\hbox{E}\kern-.125emX}}

\newcommand\blfootnote[1]{%
  \begingroup
  \renewcommand\thefootnote{}\footnote{#1}%
  \addtocounter{footnote}{-1}%
  \endgroup
}

\begin{document}

\title{A Comparison of Super-Resolution and Nearest Neighbors Interpolation Applied to Object Detection on Satellite Data\\
}

\author{\IEEEauthorblockN{Evan Koester}
\IEEEauthorblockA{\textit{Systems and Technology Research} \\
\textit Woburn, Massachusetts, USA \\
evan.koester@stresearch.com}
\and
\IEEEauthorblockN{Cem \c{S}afak \c{S}ahin}
\IEEEauthorblockA{\textit{Systems and Technology Research} \\
\textit Woburn, Massachusetts, USA \\
cem.sahin@stresearch.com}
}

\maketitle

\begin{abstract}
As Super-Resolution (SR) has matured as a research topic, it has been applied to additional topics beyond image reconstruction. In particular, combining classification or object detection tasks with a super-resolution preprocessing stage has yielded improvements in accuracy especially with objects that are small relative to the scene. While SR has shown promise, a study comparing SR and naive upscaling methods such as Nearest Neighbors (NN) interpolation when applied as a preprocessing step for object detection has not been performed. We apply the topic to satellite data and compare the Multi-scale Deep Super-Resolution (MDSR) system to NN on the xView challenge dataset. To do so, we propose a pipeline for processing satellite data that combines multi-stage image tiling and upscaling, the YOLOv2 object detection architecture, and label stitching. We compare the effects of training models using an upscaling factor of 4,  upscaling images from 30cm Ground Sample Distance (GSD) to an effective GSD of 7.5cm. Upscaling by this factor significantly improves detection results, increasing Average Precision (AP) of a generalized vehicle class by 23 percent. We demonstrate that while SR produces upscaled images that are more visually pleasing than their NN counterparts, object detection networks see little difference in accuracy with images upsampled using NN obtaining nearly identical results to the MDSRx4 enhanced images with a difference of 0.0002 AP between the two methods.

\end{abstract}


\section{Introduction}

\blfootnote{Systems \& Technology Research 2019. All Rights Reserved.}

With machine vision and deep Convolutional Neural Networks (CNNs) being applied to novel problems and data, there has been rapid growth in both network architecture advancement and the applications with which these networks are being applied. However, in most classification and object detection applications the image data is such that objects of interest are large with respect to the scene. This can be observed in the most popular public benchmark datasets ImageNet, VOC, COCO, and CIFAR [1][2][3][4].  While these datasets and their respective challenges have continued to produce incrementally advancing network architectures including SqueezeNets, Squeeze-and-Excitation Networks, and Faster R-CNN, an open challenge remains in data where objects are small relative to a larger scene [5][6][7]. This holds true for satellite data with DigitalGlobe's WorldView-3 satellite representing each pixel as a 30cm\textsuperscript{2} area. 

In these scenes, objects such as cars are often 13x7 pixels or less and are in scenes larger than 3000x3000 in size. These large scenes require preprocessing to be used in modern object detection networks that includes tiling the original scenes into smaller components for training and validation. To add to this, objects such as vehicles are often located close together in areas such as parking lots and busy roadways, making the boundaries between vehicles difficult to perceive in satellite scenes. A lack of publicly available labeled data has also hindered exploration into this application space, with only the xView Challenge Dataset having satellite captured imagery with labeled objects such as vehicles [8]. Other aerial datasets such as Classification of Fine-Grained features in Aerial images (COFGA), A large-scale Dataset for Object Detection in Aerial images (DOTA), and Cars Overhead With Context (COWC) have similar object classes, but exist at a lower Ground Sample Distance (GSD) making them easier to obtain good object detection results but limiting real-world applications [9][10][11].

Given the challenges of applying CNNs to satellite data, performing upscaling as a preprocessing step is essential to achieving good performance in accurately detecting objects. Modern advances in deep learning have resulted in a number of advanced architectures to perform upscaling that train a network on a low-resolution image and validate it versus a high-resolution copy. Despite a growing body of literature on the subject, the application of Super-Resolution (SR) to object detection and classification problems has been largely unexplored and essential comparisons between SR and naive upscaling such as Nearest Neighbors (NN) interpolation have not been documented in the literature. While SR networks show promise as a preprocessing step for object detection in satellite imagery, they also add a significant computational cost due to their deep networks containing millions of parameters that must be correctly trained. Unlike SR, NN remains one of the most basic methods of upscaling and performs interpolation by taking a neighboring pixel and assuming its value, creating a piece-wise step function approximation with little computational cost.

In this paper, we present novel contributions to the growing literature in object detection and super-resolution. Through a study of SR and NN upscaling in training and validating an object detection network applied to satellite imagery, we aim to determine the benefits of application-driven SR to offer a quantitative comparison of SR to basic interpolation methods as applied to object detection tasks. To achieve this, we propose a novel multi-stage pipeline to tile, upscale, and further tile pan-sharpened WorldView-3 satellite images into resolution-enhanced components. In doing so, we solve a documented issue with the You Only Look Once (YOLO) architecture in overcoming its challenge in accurately identifying large clusters of small objects [12]. We compare a model trained on super-resolved data to models trained on images of native resolution, NN upscaled by a factor of two, and NN upscaled by a factor of four. In doing so, we are able to evaluate the value of SR in real-world machine vision applications and whether it shows enough benefit to outweigh the large computational costs associated with a platform in which multiple deep learning architectures coexist.

\section{Related Work}

\subsection{Object Detection Architectures}
The application of object detection networks to satellite images presents a unique challenge in that satellite images are extremely large, making single objects in a scene difficult to localize and detect. While images of this kind require pre-processing to reduce the image input size to the object detection network, the detection framework chosen plays a large role in the accuracy and computational cost of performing object detection in this domain.

Modern object detection frameworks are typically split into single-stage and two-stage architectures, with the most popular of the first category including SSD, ResNet, and You Only Look Once (YOLO) [13][14][12]. These networks rely on a combination of anchor box and sliding window methods to perform region proposal and feature extraction from a scene simultaneously. A significant advantage of single-stage models is that they are designed to perform object detection quickly, with YOLO obtaining near-real time detection speeds. In the satellite domain, this translates to significant savings in detection time and computational cost.

Faster R-CNN and Mask R-CNN are common two-stage architectures that separate the region proposal and feature extraction components into separate stages and share results between stages to reduce computational load [7][15]. While these networks are more computationally expensive, they had the advantage of performance when compared to their single-stage counterparts. However advances in single-stage networks have been shown to outperform two-stage networks while performing detection significantly faster [16].

The most recent advances in object detection architectures have centered around improving the loss function that these networks use to optimize their weights. Of note are advances such as ADAM, Focal Loss, Predefined Evenly-Distributed Class Centroids Loss, and Reduced Focal Loss [17][18][19][20]. Reduced Focal Loss was used by the winners of the xView object detection in satellite imagery competition.

\subsection{Super Resolution and Overhead Image Applications}

SR networks have demonstrated the remarkable capability to upscale images and provide greatly improved visual perceptibility to humans compared to traditional interpolation methods such as NN or bicubic interpolation [21]. These networks train on low-resolution images to learn a parameter space to be able to approximate an upscaled version of the image that closely matches a held-out high-resolution copy. A variety of SR networks exist, including Super Resolution Generative Adversarial Network (SRGAN), Enhanced Deep Super Resolution (EDSR), Deep Back Projection Networks (DBPN), Super Resolution DenseNets, and Deep Laplacian Pyramid Networks (DLPN) [21][22][23][24][25]. While each of these networks has a different architecture, all are attempting the problem of single-image super-resolution on common SR datasets such as Set14, B100, and DIV2K[26][27][28].

Much of the research in SR has focused on comparing different SR architectures, usually using a Picture Signal-To-Noise Ratio (PSNR) metric to evaluate how closely an architecture can reconstruct the true high-resolution image from a low-resolution image. For this reason, these networks are rarely evaluated on the basis of computational cost, though work by Shi et. al. attempts to perform SR upscaling in near real-time using a sub-pixel CNN architecture [29]. Research into the applications of SR as a preprocessing stage for object detection and classification is much less studied, though initial studies have shown that the use of SR as a preprocessing step can yield significant improvements to the detection and classification of small objects within larger images versus detecting the objects at their native resolution [30].

For overhead images, super-resolution has been shown to both improve images visually and enhance object detector performance. Bosch investigated SR for overhead imagery that demonstrated visibly improved results on low-resolution images of airplanes [31]. This paper compared the PSNR of multiple SR networks to offer a comparison in the overhead imagery domain. While the SR enhanced images are visually compelling, this paper fails to apply the output super-resolved images to any application.

However, PSNR does not necessarily reflect improvements in applications to machine vision tasks such as image classification or object detection. Haris demonstrated this by attaching an SR network to an object detector's loss function and optimizing the SR network to minimize object detection loss [32]. Using early stopping, this would end training when the object detector reached peak performance. In doing so, Haris demonstrated that the best object detection performance occurs a lower PSNR than the maximum achievable PSNR. 

In determining the utility of SR in object detection, Shermeyer and Etten simulated various GSDs on the xView satellite image dataset and applied SR architectures to compare object detection performance across a variety of resolutions and models [30]. This study was the first of its kind to link object detection and SR on satellite imagery and demonstrated that SR provides significant enhancement in object detection on small objects such as vehicles and boats. This is echoed by Ferdous who applied SR and an SSD object detection model to the task of vehicle detection on the VEDAI aerial dataset [33]. While both of these studies showed that SR offers improvements to object detection, they also neglected to compare the effects of naive upsampling methods such as NN or bicubic interpolation on object detection performance. 

\section{Experimental Setup}

\subsection{The xView Dataset}\label{AA}
To train and validate our models using super-resolution and nearest neighbors interpolation, the xView challenge dataset was used. This dataset is composed of 1127 unique images captured using the DigitalGlobe World-View 3 satellite. Each image is captured at a GSD of 30 centimeters, causing small objects such as cars to be represented by a grid of approximately 13x7 pixels with few meaningful features for feature extraction. Because of this, the xView dataset provides great opportunity for testing the ability of object detectors to detect small objects such as vehicles. Approximately one million annotated objects are present in the dataset. While 1127 images exist in this dataset, labels only exist for 846 images. This is because the xView dataset is a competition dataset, with the organizers withholding the labels of 281 images to evaluate competition entries. 

\begin{figure}[h]
    \centering
    \mbox{
    \begin{subfigure}{0.32\columnwidth}
            \includegraphics[width=\columnwidth]{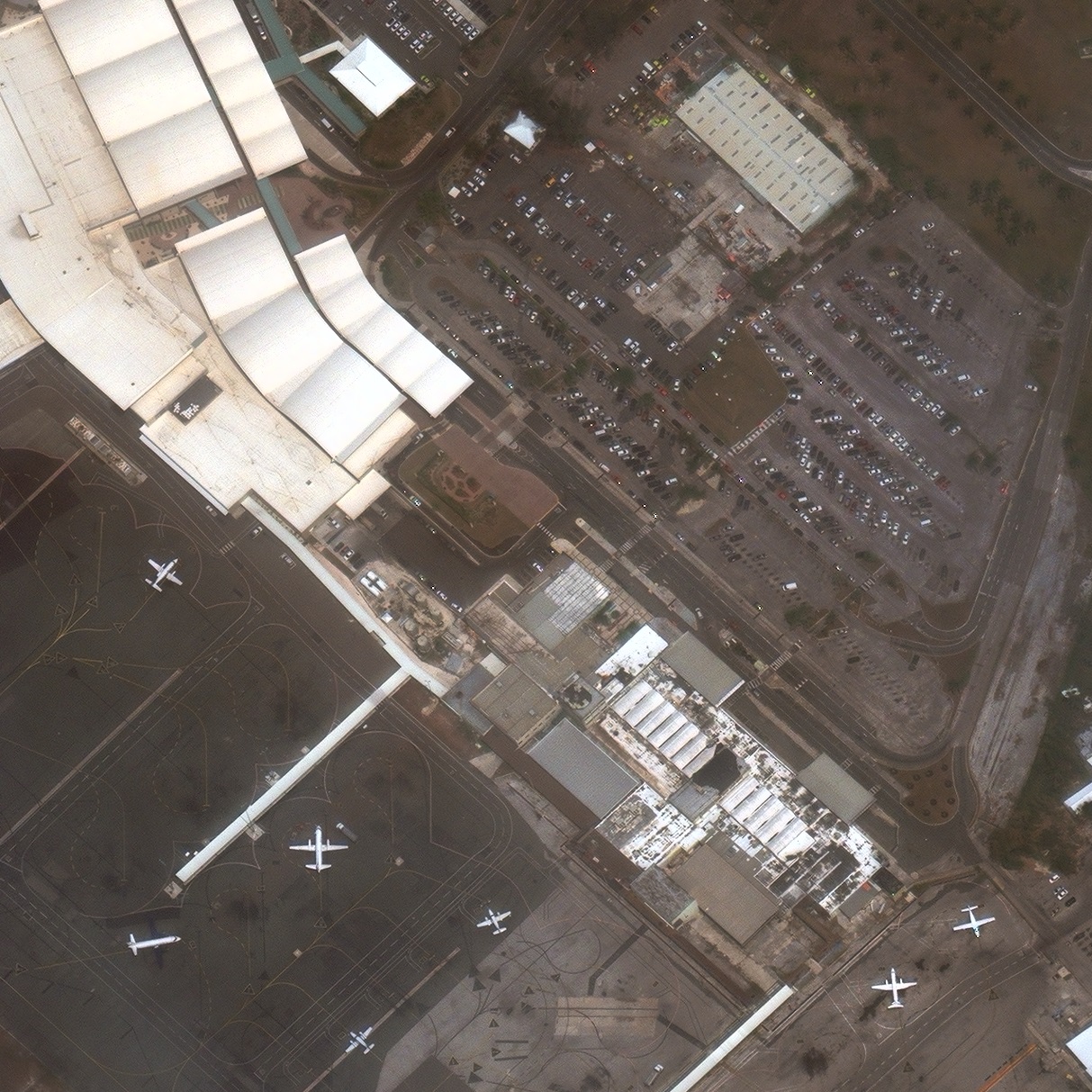}
            \caption{}
    \end{subfigure}
    \begin{subfigure}{0.32\columnwidth}
            \includegraphics[width=\columnwidth]{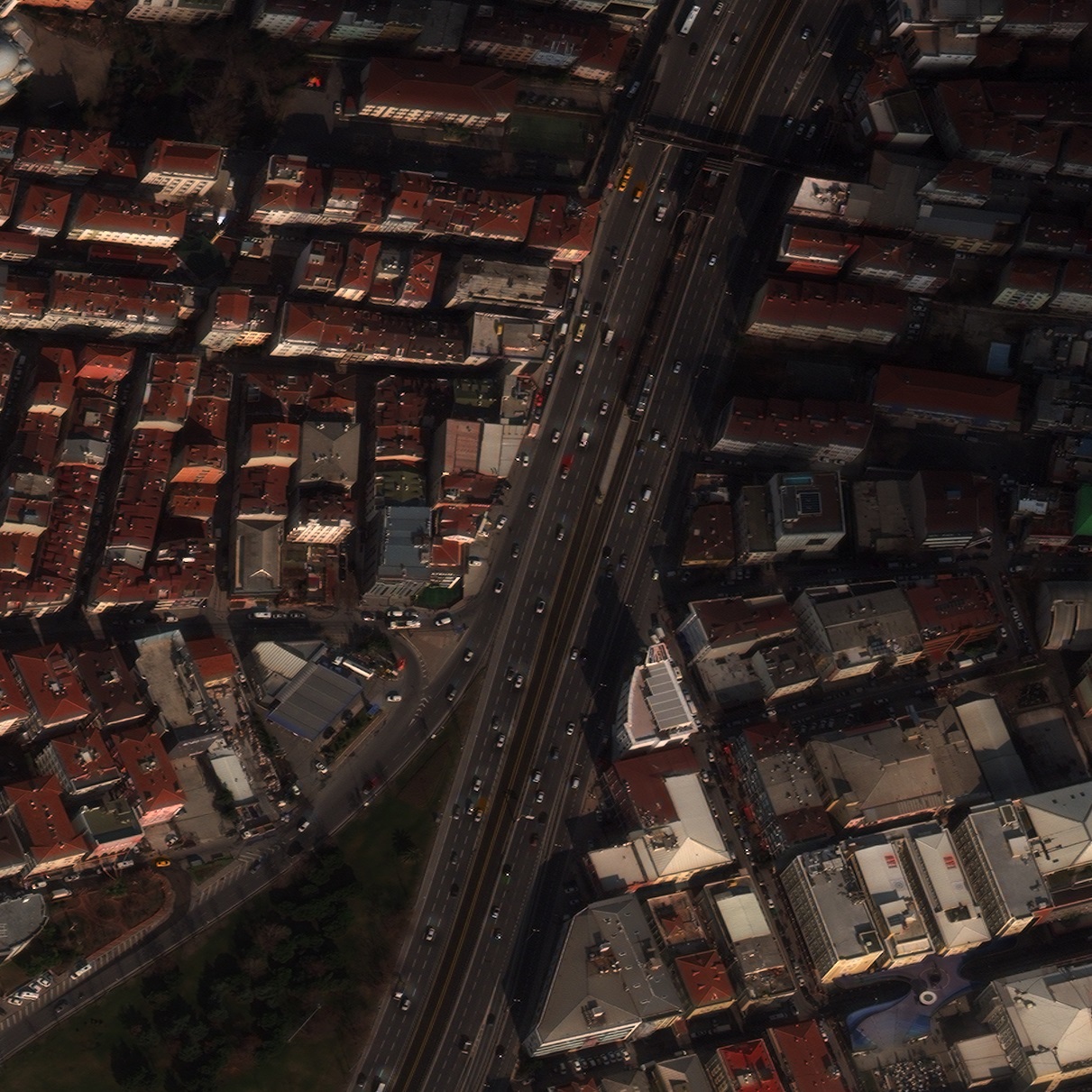}
            \caption{}
    \end{subfigure}

    \begin{subfigure}{0.32\columnwidth}
            \includegraphics[width=\columnwidth]{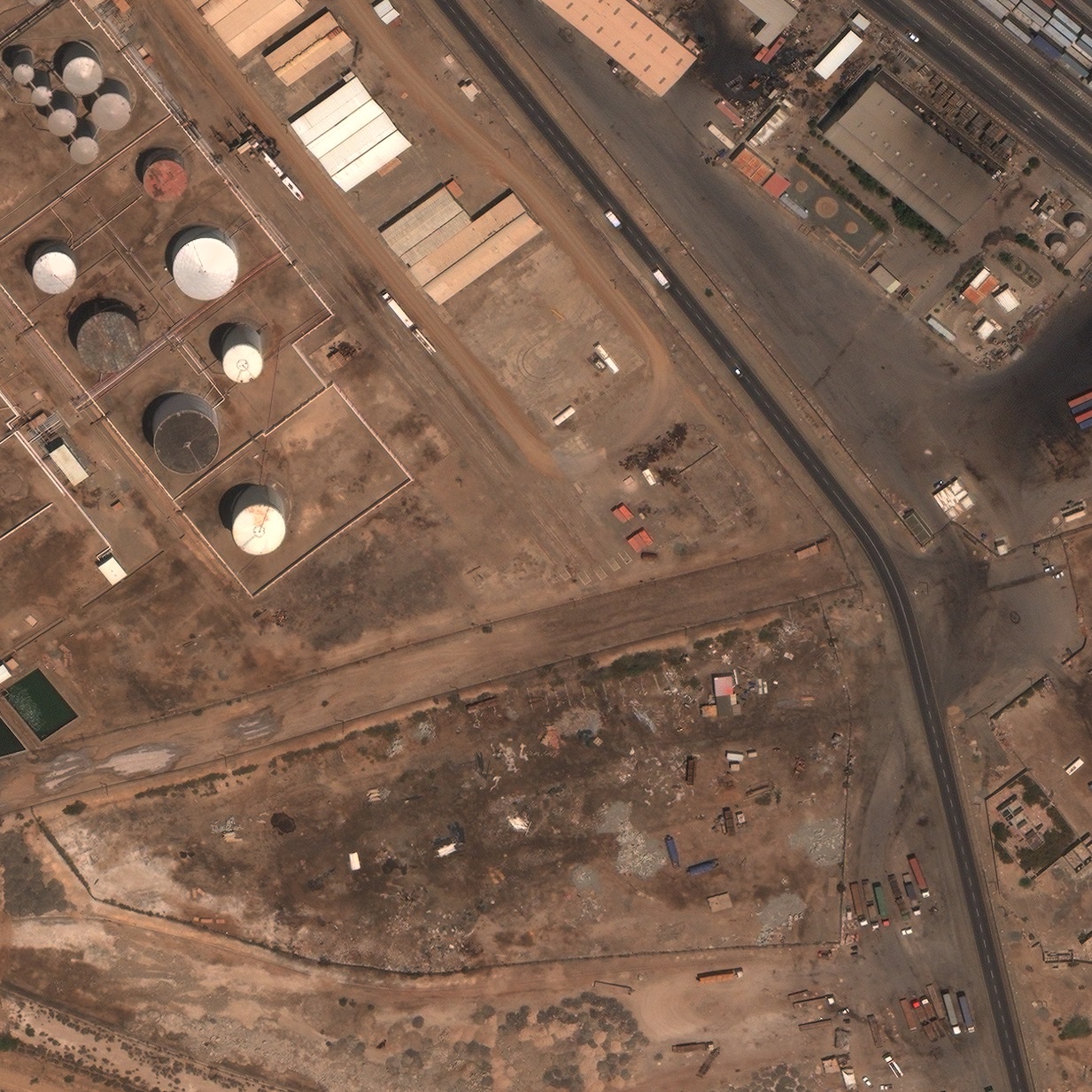}
            \caption{}
    \end{subfigure}
    }
    \caption{Example xView images including scenes such as airports (a), cities (b), and industrial areas (c)}\label{fig:xview}
\end{figure}

One highlight of the xView dataset is its range of unique labeled classes with objects such as Fixed-Wing Aircraft, Tugboat, Locomotive and Passenger Vehicle. In total, 60 classes are present with many being fine-grained classes of general object types. An example of this can be evidenced in labeled objects for \textit{Truck}, \textit{Cargo Truck}, \textit{Haul Truck} and \textit{Dump Truck}, all representing a general class for Truck. As such, we merge a total of 22 classes together to form a general vehicle class. To encourage class discrimination in training, we also train the model to recognize an airplane class and a helicopter class, though our class of interest is solely vehicles. The airplane class is represented by merging the \textit{Fixed-wing Aircraft}, \textit{Small Aircraft}, and \textit{Cargo Plane} classes.

We split the 846 labeled images into a training set of 676 images and a validation set of 170 images for an 80/20 split. We evaluate our models using an Average Precision (AP) metric across the \textit(vehicle) class and present Precision-Recall (PR) curves and quantitative images for each model.

\subsection{Object Detection and Super Resolution Models}

While many methods exist for SR, we leverage the ensemble extension of the state-of-the-art architecture of EDSR, Multi-Scale Deep Super-Resolution (MDSR). The MDSR network has the added advantage of performing 2x, 3x, and 4x upscaling in one model, allowing for us to expand upon the experiments detailed in this paper. This method combines residual learning techniques with an approach that increases the width of the network and reducing the depth. We implement the PyTorch version of EDSR/MDSR and use the MDSRx4 upscaling to compare against a NN upscaling approach. We can see in Fig. 2 that MDSR produces a more visually pleasing upscale than the NN interpolation approach. This suggests that the MDSR network succeeds in producing an upscale that has a better PSNR than NN. We leverage the pre-trained MDSR weights, as we have found the PSNR of these weights are comparable to those of a fully trained MDSR network using low-resolution images decimated from the original xView dataset.

\begin{figure}[h]
    \centering
    \mbox{
    \begin{subfigure}{0.32\columnwidth}
            \includegraphics[width=\columnwidth]{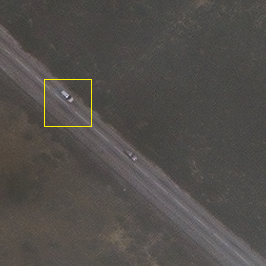}
            \caption{}
    \end{subfigure}
    \begin{subfigure}{0.32\columnwidth}
            \includegraphics[width=\columnwidth]{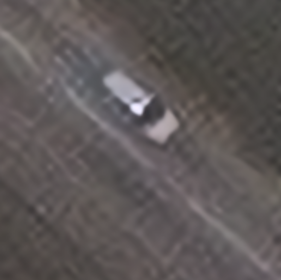}
            \caption{}
    \end{subfigure}

    \begin{subfigure}{0.32\columnwidth}
            \includegraphics[width=\columnwidth]{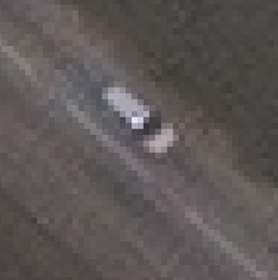}
            \caption{}
    \end{subfigure}
    }
    \caption{Upscaling a Single xView Object. We extracted a 48x48 cut from the original scene (a) to perform upscaling by a factor of 4 with MDSR (b) and Nearest Neighbors (c)}\label{fig:car}
\end{figure}

To perform our object detection experiments, we leverage the PyTorch implementation of the You Only Look Once (YOLO) v2 architecture. Satellite imagery covers extremely large areas, making fast computation of objects a requirement. This single-stage architecture performs scene detection significantly faster than more complex two-stage networks such as Mask R-CNN while achieving similar performance.

We train our YOLOv2 model on two Nvidia Titan Xp graphics cards with a batch size of 32 and subdivision size of 8. We use a learn rate of 0.00025 and momentum of 0.9. The official Darknet-19 pre-trained convolutional weights pre-trained on the ImageNet dataset are used to initialize our model. During training, we include data augmentation to add random adjustments to hue, saturation, and exposure to make our models robust to variation in color and lighting. Random translation, scale, and jitter are also used in training to increase the robustness of the model. We use a single Nvidia Titan Xp for performing object detection and evaluating performance of the trained model. To obtain our Average Precision metric we use an IOU threshold of 0.5 and sweep across bounding box confidence thresholds to obtain precision and recall values across all confidence scores from 0.01 to 0.9.

\subsection{A Custom Pipeline for Object Detection on Small Objects within Satellite Scenes}
To obtain strong performance on satellite imagery, we propose a multi-stage preprocessing pipeline that combines image tiling and upscaling. Images in the xView dataset are extremely large, with each image approximately 4000x3000 in size. Given that the YOLOv2 network resizes all input images to 416x416 to ensure a 13x13 output feature map, the original xView images must be tiled into smaller images to retain objects. Without this step, extremely small objects such as vehicles would be resized to a size of less than 4x4.

Our object detection pipeline is tuned for satellite imagery in that it performs two tiling stages combined with an upscaling step. For each image in the xView dataset, successive 208x208 tiles are cut from the image with an overlap between tiles of 50 pixels. These are fed into an upscaling stage, which enlarges each tile by a factor of 4 to produce an 832x832 cut. These upscaled images are then applied to a second stage tiler which takes 416x416 cuts with an overlap of 50 pixels. The tiling scheme is such that successive tiles are taken top to bottom until the edge of the image is reached, at which point the tiler returns to the top of the image and shifted to the right. This causes significant overlap at the edges of an image, though it is not typically noticeable for large scenes. In the second tiling stage, this significant overlap becomes an added benefit to performance by allowing YOLOv2 to have a second chance at detecting objects within a scene, often correctly detecting objects in one overlapping tile that had been missed in another. The resulting detections are then stitched back together to return bounding box locations with respect to the original scene. 

\begin{figure}[h]
	\includegraphics[width=\linewidth]{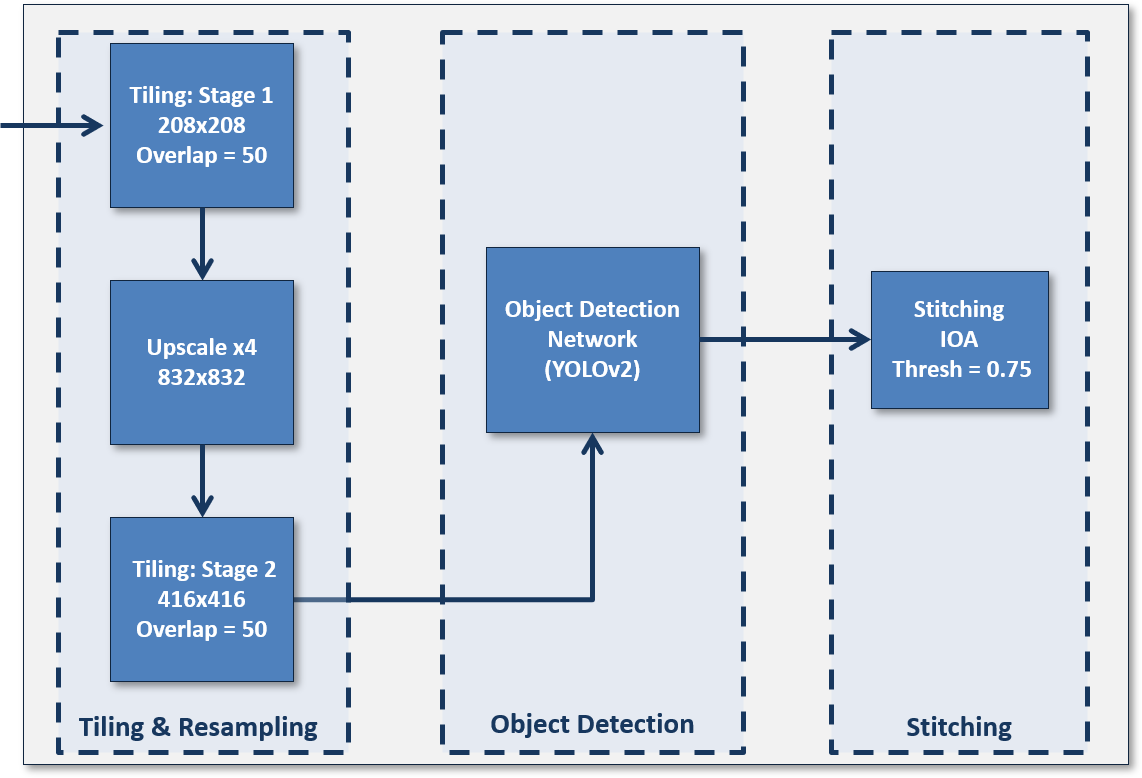}
    \caption{Object Detection Pipeline using multi-stage tiling, upscaling, YOLO9000 and stitching}
    \label{fig:PR1}
\end{figure}

To eliminate overlapping bounding boxes that occur in overlapping tile scenes, we use an Intersection over Area (IOA) metric. 

\begin{equation}
IOA = \frac{Intersection_{x} + Intersection_{y}}{min(Box1_{area},Box2_{area})}
\end{equation}

where

\begin{equation}
\begin{split}
Intersection_{x}=min(Box1_{x_{max}}, Box2_{x_{max}}) \\
 - max(Box1_{x_{min}}, Box2_{x_{min}})
\end{split}
\end{equation}

\begin{equation}
\begin{split}
Intersection_{y}=min(Box1_{y_{max}}, Box2_{y_{max}}) \\
 - max(Box1_{y_{min}}, Box2_{y_{min}})
\end{split}
\end{equation}

and box area is

\begin{equation}
Box Area = (x_{max}-x_{min})*(y_{max}-y_{min})
\end{equation}

Bounding boxes are sorted by YOLOv2's confidence score, and then compared to others that are in its vicinity. We set the threshold so that if a box overlaps  another with an intersectional area greater than 75 percent of the area of the bounding box, the box with the lower confidence score is removed. Our use of an IOA metric differs from many standard approaches that call for an Intersection-Over-Union (IOU), as we have found that IOU with an appropriate threshold value does not perform as well as IOA in merging repeated bounding boxes while retaining the bounding boxes of objects located close together, such as trucks, especially when these objects are positioned at an angle.

\section{Results}
In comparing object detection results on vehicles in the xView dataset, it is clear that tiling and upscaling plays a significant role in improving model performance on satellite imagery. As demonstrated in Table 1 and Fig. 4, NN interpolation by a factor of 2 demonstrates an improvement of 3.4\% in AP compared to the tiled images without any upscaling. Using our proposed pipeline and upscaling by a factor of 4 in combination with multi-stage tiling increases performance by 22.8\% compared to the native image GSD to reach 0.683 AP.

\begin{table}[h!]
\renewcommand{\arraystretch}{1.3}
\caption{xView Vehicle Detection AP Scores by Method}
\label{tab:example}
\centering
\begin{tabular}{c|c}
    \hline
    Method & AP Score\\
	\hline
	\hline    
    1-Stage Tiling & 0.4548\\
    \hline
    1-Stage Tiling, NNx2 & 0.4889\\
	\hline    
    2-Stage Tiling, MDSRx4 & 0.6833\\
	\hline    
    2-Stage Tiling, NNx4 & \textbf{0.6835}\\
    \hline

\end{tabular}
\end{table}

There were hardly any differences between MDSR and NN upscaling methods, with NNx4 outperforming MDSRx4 by 0.02\% in AP. Indeed when viewing the Precition-Recall curve in Fig. 4, the MDSRx4 and NNx4 curves are overlapping, suggesting that the object detector is not extracting unique features from the SR processed images that were not present in the NNx4 images. 

\begin{figure}[h]
	\includegraphics[width=\linewidth]{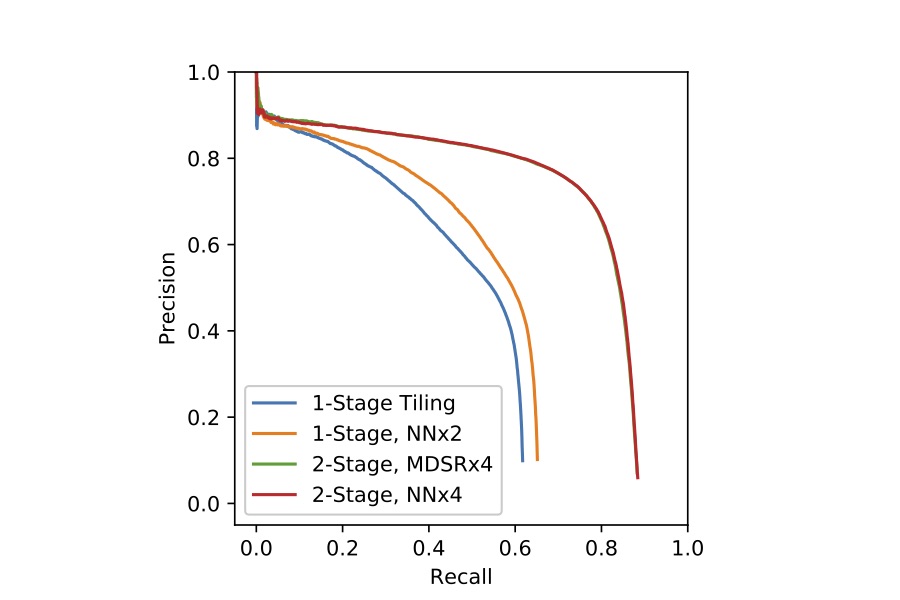}
    \caption{PR Curve demonstrating the effects of 1-Stage tiling and 2-Stage tiling with different upscaling methods}
    \label{fig:PR1}
\end{figure}

\begin{figure*}[ht]
    \centering
    \mbox{
    \begin{subfigure}{0.45\columnwidth}
            \includegraphics[width=\columnwidth]{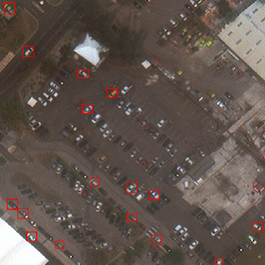}
%            \caption{1-Stage Tiling}
    \end{subfigure}
    \begin{subfigure}{0.45\columnwidth}
            \includegraphics[width=\columnwidth]{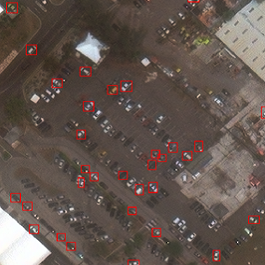}
%            \caption{1-Stage Tiling, NNx2}
    \end{subfigure}

    \begin{subfigure}{0.45\columnwidth}
            \includegraphics[width=\columnwidth]{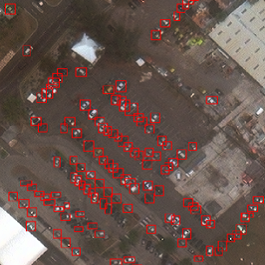}
%            \caption{2-Stage Tiling, MDSRx4}
    \end{subfigure}
    \begin{subfigure}{0.45\columnwidth}
            \includegraphics[width=\columnwidth]{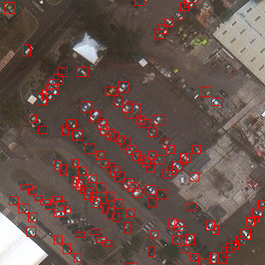}
%            \caption{2-Stage Tiling, NNx4}
    \end{subfigure}
    }
%    \caption{Object Detection Results on a Parking Lot Scene}\label{fig:ABCD}
\end{figure*}

\begin{figure*}[ht]
    \centering
    \mbox{
    \begin{subfigure}{0.45\columnwidth}
            \includegraphics[width=\columnwidth]{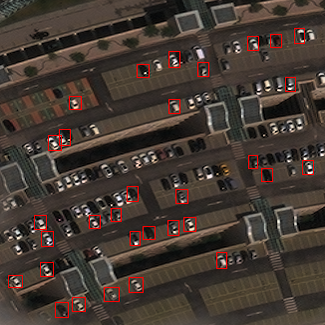}
%            \caption{1-Stage Tiling}
    \end{subfigure}
    \begin{subfigure}{0.45\columnwidth}
            \includegraphics[width=\columnwidth]{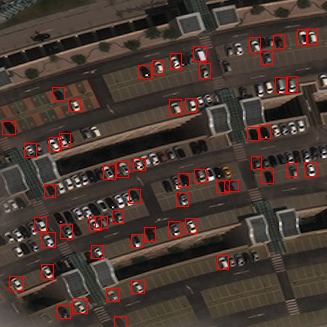}
%            \caption{1-Stage Tiling, NNx2}
    \end{subfigure}

    \begin{subfigure}{0.45\columnwidth}
            \includegraphics[width=\columnwidth]{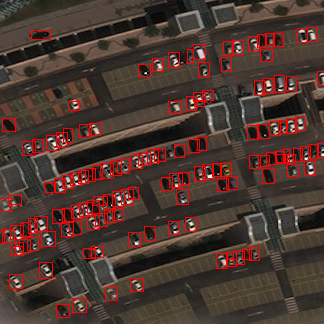}
%            \caption{2-Stage Tiling, MDSRx4}
    \end{subfigure}
    \begin{subfigure}{0.45\columnwidth}
            \includegraphics[width=\columnwidth]{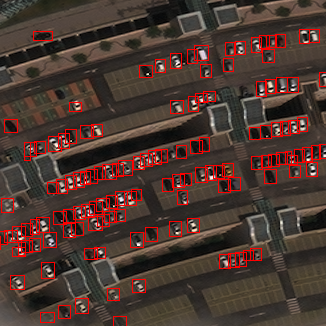}
%            \caption{2-Stage Tiling, NNx4}
    \end{subfigure}
    }
%    \caption{Object Detection Results on a Parking Lot Scene}\label{fig:ABCD}
\end{figure*}

\setcounter{figure}{4}
\begin{figure*}[ht]
    \centering
    \mbox{
    \begin{subfigure}{0.45\columnwidth}
            \includegraphics[width=\columnwidth]{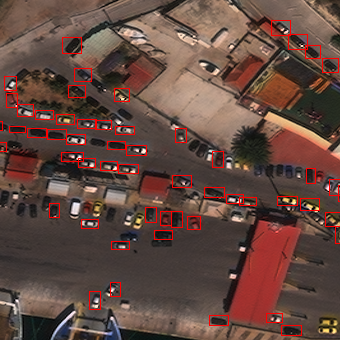}
            \caption{}
    \end{subfigure}
    \begin{subfigure}{0.45\columnwidth}
            \includegraphics[width=\columnwidth]{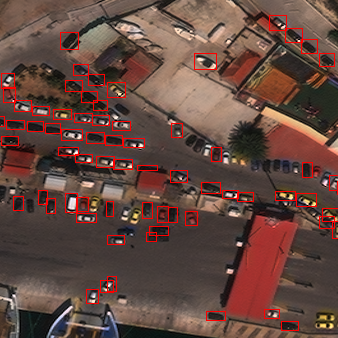}
            \caption{}
    \end{subfigure}

    \begin{subfigure}{0.45\columnwidth}
            \includegraphics[width=\columnwidth]{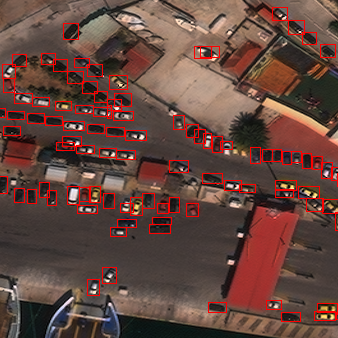}
            \caption{}
    \end{subfigure}
    \begin{subfigure}{0.45\columnwidth}
            \includegraphics[width=\columnwidth]{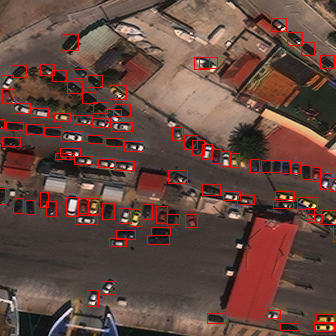}
            \caption{}
    \end{subfigure}
    }
    \caption{Object Detection Results on a Parking Lot Scene with column (a) showing a 1-stage tiling schema without upsampling, column (b) a 1-stage tiling with a NNx2 upscaling, column (c) demonstrating our proposed 2-stage tiling and upscaling pipeline with MDSRx4, and column (d) 2-stage tiling with NNx4 upscaling}\label{fig:exampleresults}
\end{figure*}

Given that the YOLOv2 model uses weights pretrained on ImageNet which are then refined on objects in the xView dataset, we suspect that the benefit we see from upscaling objects with MDSR and NN is largely due to the objects better matching ImageNet objects with regards to size in the scene. As a result, upscaling is used to aid the feature extraction process that is learned from the pretraining. Because the features are limited in satellite scenes, it is likely the network is extrapolating features from the limited geometries present in these small objects as opposed to any unique features brought about by advanced upscaling methods such as MDSR.

We note that our proposed object detection pipeline performs extremely well in scenes of low object density, as demonstrated in Fig. 4, and that this holds true for both MDSRx4 and NNx4 upscaled scenes. In crowded scenes such as roads, highways, and parking lots, our pipeline performs well  though a dip in performance is noted as the boundaries between clusters of cars becomes difficult to perceive. Given that dense clusters of small objects are a documented challenge for the YOLO object detection architecture, our pipeline's ability to overcome this network shortcoming and maintain high accuracy and fast inference speeds is a novel improvement towards applying the YOLO architecture to real-world applications.

\section{Conclusion}
In this paper we performed a novel comparison between a modern state-of-the-art SR network to a common naive interpolation method for preprocessing imagery as applied to object detection. We implemented the MDSR network for upscaling with a factor of 4, as well as a NN interpolation method, and used these upscaled images to train and validate multiple models of the YOLOv2 object detection architecture. In addition, we proposed a novel pipeline that uses a combination of multi-stage tiling and upsampling with the YOLOv2 architecture to accurately identify vehicles within large satellite scenes. In doing so, we add to the growing literature of work in object detection in overhead imagery and super-resolution. 

We demonstrated that multi-stage tiling and upscaling produce significant improvements of very small objects within large scenes, with images upscaled by a factor of 4 demonstrating a 22.8\% improvement to the single-stage tiled images at the native GSD found in the xView dataset. However, we conclude that there are minimal differences between performance with object detectors trained using NN and SR methods. Despite producing upscaled images that are closer to photo-realism than other interpolation algorithms, SR does not appear to yield any significant advantage when used in conjunction with machine vision architectures for object detection. This calls into question the value of SR for practical applications and products given that SR has a significantly higher computational cost than NN.

\end{document}